%% file: main_artigo.tex
\documentclass[runningheads]{llncs}

\usepackage[T1]{fontenc}
\usepackage{graphicx}
\usepackage{float}
\usepackage{amsmath}
\usepackage{cleveref}
\usepackage{multirow}
\usepackage[table,xcdraw]{xcolor}
\usepackage{adjustbox}

\newcommand{\beginsupplement}{%
        \setcounter{table}{0}
        \renewcommand{\thetable}{S\arabic{table}}%
        \setcounter{figure}{0}
        \renewcommand{\thefigure}{S\arabic{figure}}%
     }
     
\begin{document}
\title{XAI for Skin Cancer Detection with Prototypes and Non-Expert Supervision}

\author{Miguel Correia
\thanks{This work was supported by FCT projects LARSyS UIDB/50009/2020, 2022.07849.CEECIND, CEECIND/00326/201 and PRR projects PTSmartRetail C645440011-00000062 and Center for Responsible AI C645008882-00000055.}
\and Alceu Bissoto \and Carlos Santiago \and Catarina Barata}
\authorrunning{M. Correia et al.}

\institute{Institute  for Systems and Robotics, Instituto Superior T\'{e}cnico, Portugal}
\maketitle              
\input{Sections/abstract}
\input{Sections/New_IntroV3}
\input{Sections/ProposedApproach}
\input{Sections/ExperimentalResultsV2}

\input{Sections/conclusions}

 \bibliographystyle{splncs04}
 \bibliography{My_Bibliography_short}

\input{Sections/material_extra}
\end{document}

%% file: Sections/abstract.tex
\vspace{-0,4cm}
\begin{abstract}
Skin cancer detection through dermoscopy image analysis is a critical task. However, existing models used for this purpose often lack interpretability and reliability, raising the concern of physicians due to their black-box nature. In this paper, we propose a novel approach for the diagnosis of melanoma using an interpretable prototypical-part model. We introduce a guided supervision based on non-expert feedback through the incorporation of: 1) binary masks, obtained automatically using a segmentation network; and 2) user-refined prototypes. These two distinct information pathways aim to ensure that the learned prototypes correspond to relevant areas within the skin lesion, excluding confounding factors beyond its boundaries. Experimental results demonstrate that, even without expert supervision, our approach achieves superior performance and generalization compared to non-interpretable models. 

\keywords{Skin Cancer \and Prototypes \and Human Feedback}
\end{abstract}

%% file: Sections/New_IntroV3.tex
\section{Introduction}
Deep learning models have greatly impacted the medical field, such as in the analysis of medical images for automatic skin cancer diagnosis \cite{survey_skin_machine_learning}. 
However, many of these models are complex and difficult to understand, making them appear as black boxes. This lack of transparency creates challenges in trusting the model's decisions \cite{Jia2020}. Physicians need to rely on these models, therefore understanding their decision-making process is crucial, especially given the growing focus on eXplainable Artificial Intelligence (XAI) and increasing concerns about the use of black box deep learning models in medicine \cite{Antoniadi2021}.

XAI is a field that focuses on developing models that can provide clear explanations for their decision-making processes \cite{Roscher2020}. These explanations can be divided into two main categories: model-based and post-hoc explanations. Model-based explanations are integrated into the model during its training phase, while post-hoc explanations are generated through subsequent analysis after the model has been trained \cite{SURVEYMEDICALIMAGE}.
Various techniques exist for post-hoc explanations. One such technique is Grad-CAM, which produces saliency maps to provide local and visual explanations \cite{GradCAM}. Another technique called TCAV primarily offers textual explanations, both locally and globally \cite{Kim2017,SURVEYMEDICALIMAGE} . However, these techniques do not explicitly reveal the internal decision-making process of the model \cite{Rudin2019}. In contrast, ProtoPNet \cite{Chen2018} is an example of a model that incorporates a model-based explanation. ProtoPNet compares specific parts of an image with prototypical parts associated with certain classes, generating similarity scores that are combined to determine the final model decision. The visual explanations produced by ProtoPNet are transparent and intuitive, making it more interpretable and providing a faithful representation of the model's decision-making process.
Nevertheless,the classical version of ProtoPNet has been shown to lack diverse and clinical relevant prototypes \cite{Barnett2021,Wang2022}. This limits the interpretability of the model, as its explanations heavily rely on these prototypes. Taking as an example the skin cancer setting, when the prototypes represent artifacts like hair, air bubbles, rulers, image corners, or black edges, the explanations are compromised. Therefore, it is crucial to introduce techniques that enhance the quality of prototypes.

In this study, we integrate two prototype supervision techniques within the ProtoPNet framework for binary classification of melanoma (MEL) and nevus (NV) lesions. The first technique draws inspiration from the fine annotation method used in IAIA-BL \cite{Barnett2021}. It uses binary masks to identify the interior and boundary of the lesion, assuming that these areas have greater clinical relevance. The second technique, inspired by ProtoPDebug \cite{Bontempelli2022}, incorporates human input. Instead of binary masks, the user provides examples of valid prototypes representing the interior or boundary of the lesion. We conducted extensive experiments using various CNN backbone architectures. The ISIC 2019 dataset was used for training and validation \cite{Combalia2019,Codella2018_ISIC,Tschandl2018}, while the generalization capabilities were evaluated on the  $\text{PH}^2$ \cite{PH2_Mendonca2013} and Derm7pt \cite{Derm7pt_Kawahara2019} datasets. Our results show that our approach outperforms non-interpretable models in both performance and generalization ability. This challenges the notion that interpretable models sacrifice performance. More surprisingly, we demonstrate that even non-expert supervision has a positive impact on the quality of the prototypes and overall model performance. 

%% file: Sections/ProposedApproach.tex
\vspace{-0,3cm}
\section{Proposed Approach}
\vspace{-0,2cm}
\subsection{Model Architecture}
\begin{figure}[H]
\centering
\includegraphics[width=0.9\textwidth]{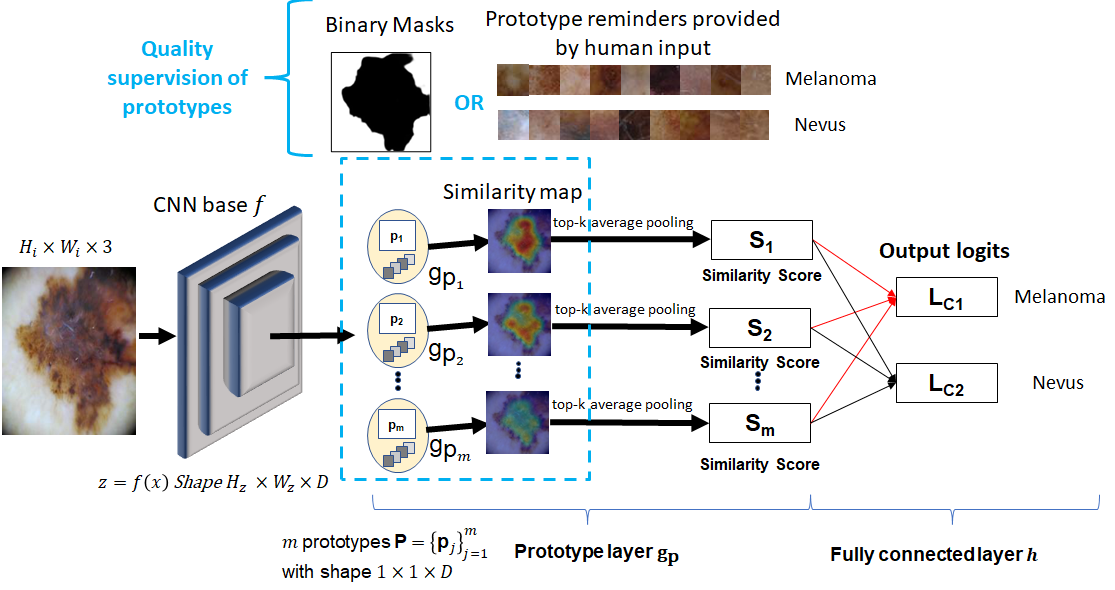}
\caption{Interpretable model for skin cancer classification of Melanoma vs Nevus. The model is based on ProtoPNet \cite{Chen2018} and enables non-expert supervision of prototypes through the use of a binary mask or human feedback.} \label{fig:Estrutura_Modelo}
\end{figure}

Our approach
\footnote{Code available at https://github.com/MiguelC23/XAI-Skin-Cancer-Detection-A-Prototype-Based-Deep-Learning-Architecture-with-Non-Expert-Supervision},
 depicted in \cref{fig:Estrutura_Modelo}, consists of three main components: 
\textit{i)} a Convolutional Neural Network (CNN) called $f$, \textit{ii)} a prototype layer denoted as $g_p$, and \textit{iii)} a fully connected layer denoted as $h$ \cite{Chen2018}. The input to the network is an image $x_i$ with dimensions $H_i \times W_i \times 3$. This image is processed by $f$, which has parameters $w_{conv}$ and outputs a feature map $z$ with dimensions
$H_z \times W_z \times D$, where each pixel in $z$ corresponds to a patch in $x_i$ and $H_z=W_z$.
The prototype layer $g_p$ learns $m$ prototypes, denoted as $P=\left\{p_j\right\}^m_{j=1}$, each one used in the corresponding prototype unit $g_{p_j}$ to compute the $L_2$ distances between $p_j$, with dimensions $1 \times 1 \times D$, and all the $H_z \times W_z$ pixels in the feature map $z$. These distances are then transformed into similarity scores, resulting in an activation map. This activation map represents the similarity scores between the prototypical part and various image regions within the unit.
\vspace{-0,4cm}
\vspace{-0,7cm}
To obtain a single value that quantifies the affinity between the prototypical part $p_j$ and the image $x_i$, top-\textit{k} average pooling is applied \cite{Barnett2021}. This pooling operation reduces the activation map to a single value, indicating how prominent the prototypical part $p_j$ is within the image. 
The output of each unit $g_{p_j}$ is calculated using
\vspace{-0,3cm}
\begin{equation}
\label{eq:output_prototype_unit_topk}
g_{p_j}(z)= \underset{\tilde{z} \in \text{patches}(z)}{\text{avg top-\textit{k}}} \log\frac{||\tilde{z}-p_j||_2+1}{||\tilde{z}-p_j||_2+\epsilon},
\end{equation}
where $\epsilon$ is a small number introduced to prevent division by zero. 
The higher the similarity score at the output of the unit $g_{p_j}$, the smaller the $L_2$ distance between a given image patch and the prototype $p_j$ in the latent space representation of the feature map $z$. Thus, a higher output value indicates a stronger presence of the prototypical part $p_j$ within the image. The user is required to define the number of classes, denoted as $K$, and the number of prototypes assigned to each class, denoted as $m_k$. The prototypes created aim to capture the essential features that distinguish each class within the classification task. In the final stage of the model, the fully connected layer $h$ employs a weight matrix $w_h$ with dimensions $K \times m$ to multiply the $m$ similarity scores obtained. This multiplication results in the score indicating the likelihood of the input image belonging to a specific class. 
The higher the similarity between the input image and the prototypes of class $k$, the greater the probability of it belonging to that class.

\subsection{Prototype Learning}
The training of our model involves three main steps: \textit{i)} training with a fixed final layer $h$, \textit{ii)} projection of prototypical parts, and \textit{iii)} training of the final layer \cite{Chen2018}. The objective is to obtain a suitable representation of the latent feature space, where features of a class are clustered around prototypical parts of that class and distant from parts of other classes. 
We will start by describing the standard training process, without non-expert supervision on prototype quality, given the training image set $D=[X,Y]=\left\{ (x_i,y_i) \right\}^n_{i=1}$. In the first step, we perform joint optimization on the convolution layers and prototypical parts of the prototype layer. 
The optimization process is defined by 
\vspace{-0,25cm}
\begin{equation}
\label{eq:SGD_TRAINING}
\underset{P,w_{conv}}{\text{min}} L_\text{P} \Leftrightarrow \underset{P,w_{conv}}{\text{min}} \frac{1}{n}  \sum_{i=1}^{n} \text{CrsEnt}(h\circ g_p \circ f(x_i), y_i) + \lambda_1 L_{\text{Clst}} + \lambda_2 L_{\text{Sep}}\quad,
\end{equation}
which includes the cross-entropy loss, the clustering loss with weight $\lambda_1$
\vspace{-0,25cm}
\begin{equation}
\label{eq:CLST_Model}
L_{\text{Clst}}=\frac{1}{n} \sum_{i=1}^{n} \underset{j:\text{class}(p_j)=y_i}{\text{min }} \frac{1}{\kappa} \sum \underset{z\in \text{patches}(f(x_i))}{\text{mink }}  ||z-p_j||_2 \quad,
\end{equation}
and the separation loss with weight $\lambda_2$
\vspace{-0,25cm}
\begin{equation}
\label{eq:SEP_Model}
L_{\text{Sep}}=-\frac{1}{n} \sum_{i=1}^{n} \underset{j:\text{class}(p_j)\neq y_i}{\text{min }} \frac{1}{\kappa} \sum \underset{z\in \text{patches}(f(x_i))}{\text{mink }}  ||z-p_j||_2\quad.
\end{equation}
The second step projects prototypes onto the closest latent parts of training images from the same class, enabling the direct association of learned prototypes as parts of images. We ensure that prototypes are projected onto different training images to promote diversity \cite{Wang2022}. In the final step, we conduct convex optimization on the parameters of the last layer while keeping the prototype layer fixed
\vspace{-0,25cm}
\begin{equation}
\label{eq:last_layer_convex_optimization}
\underset{w_h}{\text{min }} \frac{1}{n}  \sum_{i=1}^{n} \text{CrsEnt}(h\circ g_p \circ f(x_i), y_i) + \lambda_5  \sum_{k=1}^{K} \sum_{j:p_j \notin P_k}^{} |w^{ (k,j) }_h|.
\end{equation}
This optimization aims to minimize cross-entropy loss and promote sparsity in the last layer’s parameter matrix.
\subsection{Non-expert Supervision of Prototypes}
Additional terms can be added to the cost function $L_\text{P}$ in \eqref{eq:SGD_TRAINING}, to regulate and supervise the quality of the prototypes. This ensures that the prototypes represent the most relevant aspects of skin lesions and avoid confounding factors such as black borders, image corners, rulers, or hair. We explored two alternative approaches: the mask loss $L_{\text{M}}$, which is associated with lesion segmentation masks, inspired by the fine-annotation loss \cite{Barnett2021}, or the remembering loss $L_{\text{R}}$ derived from ProtoPDebug \cite{Bontempelli2022} that explores direct human-feedback on the prototypes. Both losses can be added to \eqref{eq:SGD_TRAINING}, with appropriate weights $\lambda_3$ and $\lambda_4$.

\textbf{Mask Loss ($L_{\text{M}}$):}
As mentioned previously, the primary objective of $L_{\text{M}}$ is to incorporate a supervision technique to ensure that the prototypes possibly represent more significant characteristics of skin lesions for diagnostic purposes, while minimizing the impact of confounding factors. To achieve this, a binary mask is employed to distinguish between relevant pixels (lesion, marked as 0) and non-relevant pixels (skin and other elements, marked as 1). Given the difficulty to obtain segmentation masks provided by experts at a large scale, in this work we explore the use of an automatic segmentation model \cite{SEGMENTATION_MODEL} to collect most of the binary masks. This approach has its limitations, as lesion segmentation networks still fail for more challenging cases \cite{mirikharaji2023}. Thus, it can be perceived as a proxy for binary masks provided by a non-expert. 

A crucial step of this approach involves calculating the element-wise product between the binary mask $M_i$ and the prototype activation map $\text{PAM}_{i,j}=\text{ScaleUp}(g_{p_j}(f(x_i)))$, where the ScaleUp operator resizes the activation map from dimensions $H_z \times W_z$ to dimensions $H_i \times W_i$ using a 2D adaptive average pooling technique. For a particular training image $x_i$ belonging to class $y_i$ and with corresponding mask $M_i$, and given a skin lesion prototype $p_j$ belonging to the same class $y_i$, the binary mask loss is defined by
\vspace{-0,25cm}
\begin{equation}
\label{eq:LM}
L_{\text{M}}= \sum_{i\in D}^{} 
\underset{j:\text{class}(p_j)=y_i}{\sum} ||\text{M}_i \odot \text{PAM}_{i,j} ||_2. 
\end{equation}

By performing an element-wise multiplication between $M_i$ and the prototype activation map, we obtain a map that indicates the regions in the image where the prototype shows activity in areas of low clinical significance. Consequently, the summation within the binary mask loss aims to minimize the number of prototype activations in these clinically insignificant areas, when these prototypes belong to the same class $k$ as the training image $x_i$. 
This approach encourages the training algorithm to learn prototypes that effectively capture medically relevant characteristics specific to their assigned categories in skin lesions. 

\textbf{Remembering Loss ($L_{\text{R}}$):}
The remembering loss within the ProtoPDebug framework offers a compelling alternative to $L_{\text{M}}$ \cite{Bontempelli2022}. Unlike $L_{\text{M}}$, which relies on binary masks to identify clinically relevant regions, the remembering loss allows users to provide examples of relevant prototypes that they consider characteristic or representative of each skin lesion class. These user-selected prototypes, presented as images to the model and chosen through human input, serve as valid examples that should activate the prototypes generated by the model during the training process.  
In this work, such feedback was given by a non-expert, who followed simple heuristics (\textit{e.g.}, remember prototypes inside and in the border of skin lesions, and discard prototypes that activated in artifacts, such as dark corners, or skin). However, due to its simplicity, in the future a medical expert could be included in the process. 

Let $V$ denote the set of $n_{rp}$  valid prototypes given by the user. For each $v_i \in V$, similarity scores between $v_i$ and the prototypes belonging to the same class as $v_i$ are calculated and summed. This process is repeated for all elements in $V$, and the average is taken, leading to\footnote{According to the code provided on the official GitHub repository at https://github.com/abonte/protopdebug.}

\vspace{-0,25cm}
\begin{equation}
\label{eq:LR}
L_{\text{R}} = -\frac{1}{n_{rp}} \sum_{i=1}^{n_{rp}} \quad
\underset{j:\text{class}(p_j)=\text{class}(v_i)}{\sum} 
\log \frac{||p_j-v_i||_2+1}{||p_j-v_i||_2+\epsilon} \quad.
\end{equation}
The remembering loss aims to maximize the activation of prototypes belonging to a specific class $k$ relative to the input example prototypes considered valid for that class $k$.
 In contrast to ProtoPDebug, our approach does not involve initial training and subsequent debugging rounds. We train the model from the beginning with examples of valid prototypes, chosen from previously trained models without $L_\text{R}$, making it more comparable to $L_\text{M}$.

%% file: Sections/ExperimentalResultsV2.tex
\section{Experimental Results}
\subsection{Experimental Setup}
The training process was conducted using a partition of the ISIC 2019 dataset \cite{Combalia2019,Codella2018_ISIC,Tschandl2018},  focused on the MEL and NV classes. The training set consisted of 3,611 MEL images and 10,293 NV images. The validation set included 904 MEL images and 2,575 NV images. To evaluate the generalization capabilities of the models when exposed to test sets from a different hospital domain with distinct distributions, we utilized the $\text{PH}^2$ \cite{PH2_Mendonca2013} dataset, which comprised 40 MEL images and 160 NV images, as well as the Derm7pt \cite{Derm7pt_Kawahara2019} dataset, which included 252 MEL images and 575 NV images. The models were trained using PyTorch on a NVIDIA GeForce RTX 3090. The evaluation metrics considered were balanced accuracy (BA) and the recall for each class (R-MEL and R-NV). 

Five different architectures were employed: ResNet-18, ResNet-50, EfficientNet B3, Densenet-169, and VGG-16. These architectures served as both baseline models for classification comparison and as the CNN backbone, $f$, see \cref{fig:Estrutura_Modelo}. The CNN backbones were initialized using the weights pre-trained on ImageNet. Three scenarios were considered: \textit{i}) without non-expert supervision on the quality of prototypes ($L_\text{P}$), \textit{ii}) with non-expert supervision using $L_\text{P}+L_\text{M}$, and \textit{iii}) with non-expert supervision using $L_\text{P}+L_\text{R}$. The number of prototypes remained fixed at $m=18$, with $m_k=9$ for $k \in \{0,1\}$ and $K=2$. 

The model was trained for 21 epochs, from 0 to 20, with the projection of prototypes performed at epochs 5, 10, 15, and 20. During the projection epochs, 10 iterations of optimization of the last layer are performed. It is important to note that epochs 0 to 4 serve as a warm-up phase. Similar to IAIA-BL \cite{Barnett2021}, the warm-up epochs train the parameters of the models related to the two convolution layers (add-on layers) before the prototype layer with a learning rate of $2\times10^{-3}$, while the prototypes are trained with a learning rate of $3\times10^{-3}$. In the standard epochs, the parameters related to features, add-on layers, and prototypes are updated with learning rates of $2\times10^{-4}$, $3\times10^{-3}$, and $3\times10^{-3}$, respectively, and a step size of 5. In the iterations of optimization of the last layer after the projection process, the learning rate is set to $1\times10^{-3}$. 
The optimization algorithm used was Adam. 

The cost function coefficients used were as follows: $\lambda_1=0.8$, $\lambda_2=0.08$, $\lambda_3=0.001$ ($L_\text{M}$ coefficient), $\lambda_4=0.02$ ($L_\text{R}$ coefficient), and $\lambda_5=1\times10^{-4}$ \cite{Chen2018,Barnett2021,Bontempelli2022} \footnote{Based on empirical observations, the reported coefficient values in the referenced articles were suitable for our problem and validation dataset.}.
The training batch size was set to 75. 
The input images had dimensions $H_i = W_i = 224$. 
The models were trained with the following possible values for $D$: 128, 256, 512. The output feature map $z$ of the CNN backbone had $H_z=7$ for all architectures except for VGG-16, where $H_z=14$. The models were trained with 13 possible values for the top-\textit{k} between 1 and the maximum value $H_z\times W_z$. It is important to mention that the binary masks used are 14\% partly sourced from the ISIC dataset, while 86\% were obtained using a segmentation algorithm \cite{SEGMENTATION_MODEL}. It is also important to note that in the $L_\text{P}+L_\text{R}$ approach, only the top-performing 10 configurations (for each architecture) of $D$ and top-\textit{k} were trained, based on the results obtained from the $L_\text{P}$ and $L_\text{P}+L_\text{M}$ approaches. Additionally, the number of valid prototypes provided per class and chosen by the human user is 25 in the $L_\text{P}+L_\text{R}$  approach.

\subsection{Results and Analysis}
Table \ref{tab:results} presents the top results achieved in the validation set (ISIC 2019) along with their corresponding results in the test sets.
\vspace{-0,3cm}
\input{Sections/results_tableV2}

\textbf{$L_\text{P}$ or $L_\text{P}+L_\text{M}$ vs Baseline:} Beginning with an analysis of the results on the validation set of ISIC 2019, it is evident that in 4 out of the 5 architectures utilized, $L_\text{P}$ in conjunction with the non-expert supervision introduced by $L_\text{M}$, outperformed both the $L_\text{P}$ approach and their respective baseline models.

\begin{figure}[H]
\includegraphics[width=\textwidth]{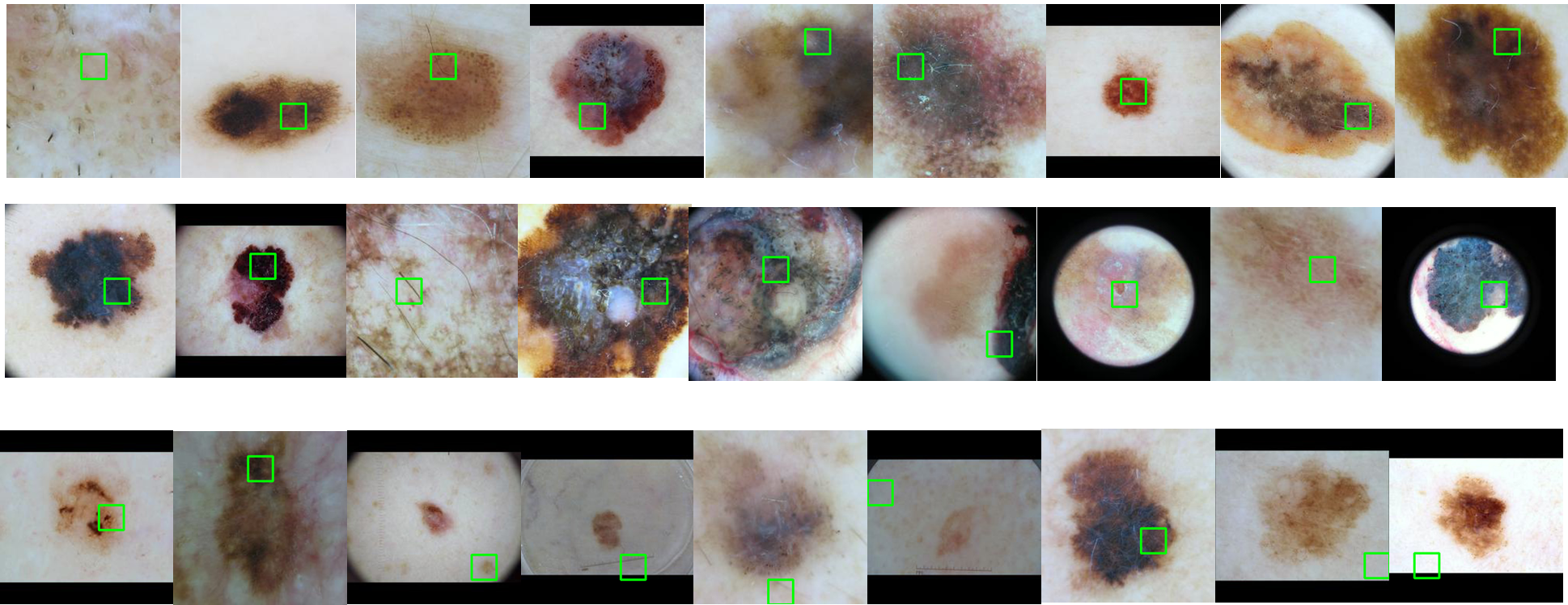}
\caption{
Melanoma prototypes obtained using EfficientNet B3 CNN backbone with the best-performing models on ISIC 2019 validation set, for different approaches: $L_\text{P}$ + $L_{\text{M}}$ (1st line), $L_\text{P}$ + $L_\text{R}$ (2nd line), and $L_\text{P}$ (3rd line).
In $L_P$, prototypes are often found near black edges, corners, instead of within the lesion, unlike the other two approaches} \label{fig:PrototypesExampleMEL}
\end{figure}

This highlights the possibility of having an interpretable model that competes
with, or even surpasses, non-interpretable models. Furthermore, by analyzing the prototypes qualitatively, see \cref{fig:PrototypesExampleMEL}, we can observe the improvement in their quality, because the prototypes depict interior components of the lesion rather than black borders and other confounding factors present in the image, thereby enhancing the reliability of explanations heavily dependent on them. 
\vspace{-0,2cm}

\begin{figure}[H]
\centering
\includegraphics[width=0.8\textwidth]{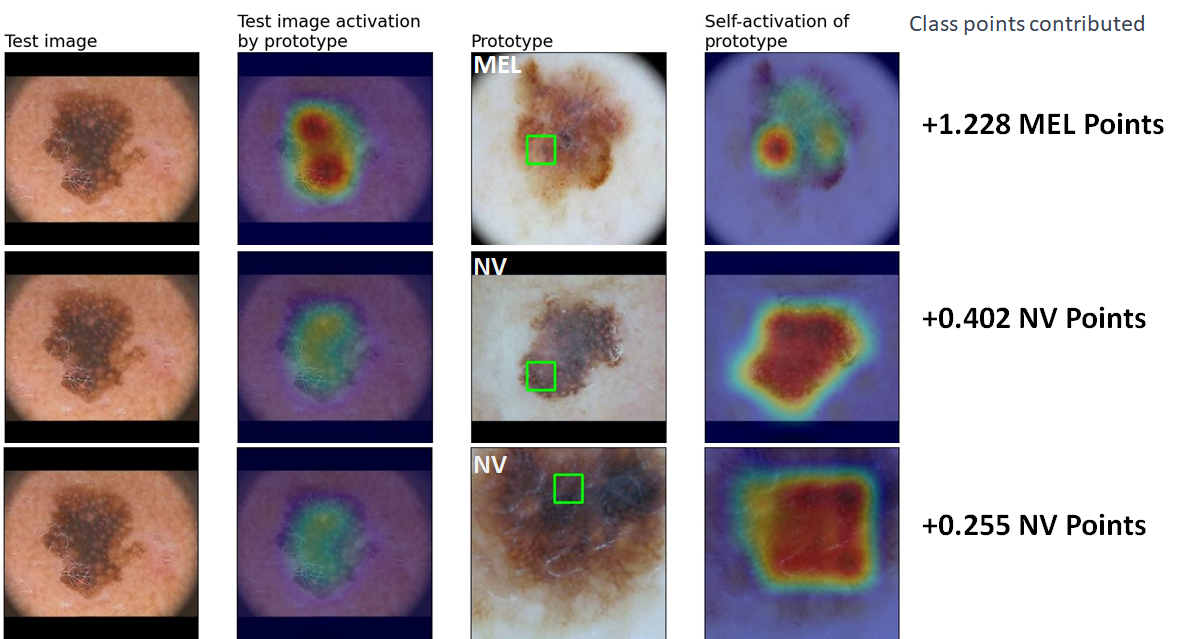}
\caption{Explanation of $L_\text{P}$ + $L_\text{M}$ approach for a test case from the $\text{PH}^2$ dataset with the top 3 activated prototypes in ResNet-50 backbone. The first prototype is melanoma, while the other two are nevus. The test image is melanoma and is correctly classified. The explanation is based on the similarity between parts of the test image and the prototypes.} \label{fig:Explanation}
\end{figure}

\vspace{-0,3cm}
\textbf{$L_\text{P}$ or $L_\text{P}+L_\text{M}$ Generalization Strength:} The $L_\text{P}$ approach outperformed non-interpretable baseline models in 3 out of 5 architectures in both $\text{PH}^2$ and Derm7pt datasets. The $L_\text{P}$ + $L_\text{M}$ approach outperformed non-interpretable baseline models in 4 out of 5 cases for $\text{PH}^2$ and 3 out of 5 cases for Derm7pt. 
Additionally, $L_\text{P}$ + $L_\text{M}$ was superior to $L_\text{P}$ in 3 out of 5 cases for $\text{PH}^2$ and 4 out of 5 cases for Derm7pt. 
It is noteworthy to mention that the Derm7pt dataset is significantly older than the other datasets, and there is a considerably larger domain shift in terms of data quality. These results demonstrate the superior generalization of our interpretable approaches compared to their non-interpretable counterparts. Furthermore, this highlights the importance of supervision to improve prototype and prevent biases from dataset artifacts, which could compromise model generalization and the explanation provided, see \cref{fig:Explanation}.

\textbf{$L_\text{P}$+$L_\text{R}$ vs Others:}
When analyzing the results, we found that using $L_\text{P}$ along with alternative supervision $L_\text{R}$ led to inferior performance on the ISIC 2019 validation set for 4 out of 5 architectures compared to $L_\text{P}$ and $L_\text{P}$ + $L_\text{M}$. This is likely because when users provide valid prototype examples, the model adjusts the prototypes to be more similar to and influenced by the user's input. As a result, the model exhibits a more restricted behavior compared to using $L_\text{M}$. Moreover, 3 out of 5 architectures performed worse than non-interpretable baseline models. However, with $L_\text{R}$, at least 80\% of the prototypes represented the interior or boundary of the skin lesion, whereas without it, the percentage was lower. $L_\text{M}$ ensured that 100\% of the prototypes met this criterion. The $L_\text{R}$ approach outperformed non-interpretable baseline models in 2 out of 5 cases for
$\text{PH}^2$  and 3 out of 5 cases for Derm7pt. However, the generalization capability of $L_\text{R}$ in these datasets, when compared to $L_\text{P}$ and $L_\text{P}$+$L_\text{M}$, was inferior. 
\vspace{-0,2cm}
\vspace{-0,3cm}
In general, it can be concluded that the $L_\text{P}$+$L_\text{M}$ approach yielded the best results in terms of performance and generalization, ensuring that the prototypes did not represent confounding factors outside the boundary of the skin lesion.

%% file: Sections/results_tableV2.tex
\begin{table}[H]
\centering
\caption{Performances with different approaches: best results on ISIC 2019 \cite{Combalia2019,Codella2018_ISIC,Tschandl2018}, along with their corresponding results in $\text{PH}^2$ \cite{PH2_Mendonca2013}, and Derm7pt \cite{Derm7pt_Kawahara2019} test sets. The best results for each architecture are highlighted in color according to the metric used.
}
\label{tab:results}
\addtolength{\tabcolsep}{-0.6pt}
\begin{tabular}{|c|c|ccc|ccc|ccc|}
\hline
 &  & \multicolumn{3}{c|}{\textbf{\begin{tabular}[c]{@{}c@{}}Best Results\\ ISIC 2019\end{tabular}}} & \multicolumn{3}{c|}{\textbf{\begin{tabular}[c]{@{}c@{}}Results\\ PH2\end{tabular}}} & \multicolumn{3}{c|}{\textbf{\begin{tabular}[c]{@{}c@{}}Results\\ Derm7pt\end{tabular}}} \\ \cline{3-11} 
\multirow{-2}{*}{\textbf{Model}} & \multirow{-2}{*}{\textbf{Approach}} & \textbf{BA} & \textbf{R-MEL} & \textbf{R-NV} & \textbf{BA} & \textbf{R-MEL} & \textbf{R-NV} & \textbf{BA} & \textbf{R-MEL} & \textbf{R-NV} \\ \hline
 & Baseline & \cellcolor[HTML]{FFFFFF}83.1 & \cellcolor[HTML]{FFFFFF}73.9 & \cellcolor[HTML]{FFFFFF}92.3 & 71.9 & 45.0 & 98.7 & 70.6 & 51.2 & 90.1 \\
 & $L_\text{P}$ & \cellcolor[HTML]{FFFFFF}85.7 & \cellcolor[HTML]{BDD7EE}79.6 & \cellcolor[HTML]{FFFFFF}91.8 & \cellcolor[HTML]{FFFF00}{\color[HTML]{000000} 78.7} & \cellcolor[HTML]{BDD7EE}60.0 & 97.5 & 73.1 & \cellcolor[HTML]{BDD7EE}58.3 & 87.8 \\
 & $L_\text{P}$+$L_\text{M}$ & \cellcolor[HTML]{FFFF00}85.8 & \cellcolor[HTML]{FFFFFF}79.3 & \cellcolor[HTML]{FFFFFF}92.3 & 78.4 & 57.5 & 99.4 & \cellcolor[HTML]{FFFF00}{\color[HTML]{000000} 73.5} & 51.6 & \cellcolor[HTML]{C6E0B4}95.5 \\
\multirow{-4}{*}{\rotatebox[origin=c]{90}{\textbf{RN-18}}} & $L_\text{P}$+$L_\text{R}$ & \cellcolor[HTML]{FFFFFF}84.8 & \cellcolor[HTML]{FFFFFF}76.0 & \cellcolor[HTML]{C6E0B4}93.6 & 77.5 & 55.0 & \cellcolor[HTML]{C6E0B4}100.0 & 72.1 & 52.8 & 91.4 \\ \hline
 & Baseline & \cellcolor[HTML]{FFFFFF}85.2 & \cellcolor[HTML]{FFFFFF}79.3 & \cellcolor[HTML]{FFFFFF}91.1 & 65.6 & 32.5 & \cellcolor[HTML]{C6E0B4}98.7 & 68.5 & 44.8 & \cellcolor[HTML]{C6E0B4}92.2 \\
 & $L_\text{P}$ & \cellcolor[HTML]{FFFFFF}84.7 & \cellcolor[HTML]{FFFFFF}79.0 & \cellcolor[HTML]{FFFFFF}90.5 & 77.2 & 57.5 & 96.9 & 71.3 & 52.0 & 90.6 \\
 & $L_\text{P}$+$L_\text{M}$ & \cellcolor[HTML]{FFFF00}{\color[HTML]{000000} 86.1} & \cellcolor[HTML]{BDD7EE}80.7 & \cellcolor[HTML]{C6E0B4}91.5 & \cellcolor[HTML]{FFFF00}86.25 & \cellcolor[HTML]{BDD7EE}75.0 & 97.5 & \cellcolor[HTML]{FFFF00}74.8 & \cellcolor[HTML]{BDD7EE}58.7 & 90.9 \\
\multirow{-4}{*}{\rotatebox[origin=c]{90}{\textbf{RN-50}}} & $L_\text{P}$+$L_\text{R}$ & \cellcolor[HTML]{FFFFFF}82.7 & \cellcolor[HTML]{FFFFFF}77.5 & \cellcolor[HTML]{FFFFFF}87.8 & 74.1 & 50.0 & 98.1 & 69.3 & 55.6 & 83.0 \\ \hline
 & Baseline & \cellcolor[HTML]{FFFFFF}83.2 & \cellcolor[HTML]{FFFFFF}79.3 & \cellcolor[HTML]{FFFFFF}87.1 & \cellcolor[HTML]{FFFF00}83.1 & \cellcolor[HTML]{BDD7EE}72.5 & 93.7 & 76.0 & 62.3 & 89.7 \\
 & $L_\text{P}$ & \cellcolor[HTML]{FFFFFF}87.3 & \cellcolor[HTML]{BDD7EE}86.1 & \cellcolor[HTML]{FFFFFF}88.5 & 78.7 & 60.0 & 97.5 & \cellcolor[HTML]{FFFF00}78.6 & \cellcolor[HTML]{BDD7EE}69.4 & 87.8 \\
 & $L_\text{P}$+$L_\text{M}$ & \cellcolor[HTML]{FFFF00}87.3 & \cellcolor[HTML]{FFFFFF}81.4 & \cellcolor[HTML]{C6E0B4}93.3 & 75.0 & 50.0 & \cellcolor[HTML]{C6E0B4}100.0 & 73.4 & 50.8 & \cellcolor[HTML]{C6E0B4}96.0 \\
\multirow{-4}{*}{\rotatebox[origin=c]{90}{\textbf{EN-B3}}} & $L_\text{P}$+$L_\text{R}$ & \cellcolor[HTML]{FFFFFF}85.5 & \cellcolor[HTML]{FFFFFF}80.4 & \cellcolor[HTML]{FFFFFF}90.7 & 78.1 & 62.5 & 93.7 & 72.8 & 54.4 & 91.3 \\ \hline
 & Baseline & \cellcolor[HTML]{FFFFFF}86.4 & \cellcolor[HTML]{FFFFFF}78.9 & \cellcolor[HTML]{C6E0B4}93.9 & 71.2 & 42.5 & \cellcolor[HTML]{C6E0B4}100.0 & \cellcolor[HTML]{FFFF00}74.8 & \cellcolor[HTML]{BDD7EE}59.5 & 90.1 \\
 & $L_\text{P}$ & \cellcolor[HTML]{FFFF00}87.5 & \cellcolor[HTML]{BDD7EE}83.0 & \cellcolor[HTML]{FFFFFF}92.1 & 73.4 & 47.5 & 99.4 & 71.6 & \cellcolor[HTML]{FFFFFF}57.1 & 86.1 \\
 & $L_\text{P}$+$L_\text{M}$ & \cellcolor[HTML]{FFFFFF}86.8 & \cellcolor[HTML]{FFFFFF}80.7 & \cellcolor[HTML]{FFFFFF}92.9 & \cellcolor[HTML]{FFFF00}75.0 & \cellcolor[HTML]{BDD7EE}52.5 & 97.5 & 72.5 & 54.4 & \cellcolor[HTML]{C6E0B4}90.6 \\
\multirow{-4}{*}{\rotatebox[origin=c]{90}{\textbf{DN-169}}} & $L_\text{P}$+$L_\text{R}$ & \cellcolor[HTML]{FFFFFF}85.6 & \cellcolor[HTML]{FFFFFF}78.6 & \cellcolor[HTML]{FFFFFF}92.5 & 69.7 & 40.0 & 99.4 & 70.2 & 51.2 & 89.2 \\ \hline
 & Baseline & \cellcolor[HTML]{FFFFFF}85.0 & \cellcolor[HTML]{BDD7EE}81.3 & \cellcolor[HTML]{FFFFFF}88.7 & 80.0 & 65.0 & 95.0 & 71.6 & 56.3 & 86.8 \\
 & $L_\text{P}$ & \cellcolor[HTML]{FFFFFF}83.9 & \cellcolor[HTML]{FFFFFF}77.4 & \cellcolor[HTML]{FFFFFF}90.3 & \cellcolor[HTML]{FFFFFF}76.6 & \cellcolor[HTML]{FFFFFF}70.0 & \cellcolor[HTML]{FFFFFF}83.1 & \cellcolor[HTML]{FFFFFF}67.9 & \cellcolor[HTML]{FFFFFF}46.8 & 89.0 \\
 & $L_\text{P}$+$L_\text{M}$ & \cellcolor[HTML]{FFFF00}85.5 & \cellcolor[HTML]{FFFFFF}79.5 & \cellcolor[HTML]{C6E0B4}91.4 & \cellcolor[HTML]{FFFF00}83.7 & \cellcolor[HTML]{BDD7EE}70.0 & \cellcolor[HTML]{C6E0B4}97.5 & \cellcolor[HTML]{FFFF00}75.1 & \cellcolor[HTML]{BDD7EE}62.7 & \cellcolor[HTML]{FFFFFF}87.5 \\
\multirow{-4}{*}{\rotatebox[origin=c]{90}{\textbf{VGG-16}}} & $L_\text{P}$+$L_\text{R}$ & \cellcolor[HTML]{FFFFFF}84.5 & 80.0 & \cellcolor[HTML]{FFFFFF}89.0 & \cellcolor[HTML]{FFFFFF}75.0 & \cellcolor[HTML]{FFFFFF}55.0 & \cellcolor[HTML]{FFFFFF}95.0 & \cellcolor[HTML]{FFFFFF}74.7 & \cellcolor[HTML]{FFFFFF}60.3 & \cellcolor[HTML]{C6E0B4}89.0 \\ \hline
\end{tabular}
\end{table}

%% file: Sections/conclusions.tex
\section{Conclusions}
This study demonstrates how our approach can be utilized as an interpretable model in a skin cancer diagnostic problem, using additional non-expert information to ensure that the prototypes represent more relevant characteristics of their respective classes. One potential source of information is the input derived directly from the selection of relevant prototypes by the user. Additionally, it shows how our model can outperform non-interpretable models in terms of performance and generalization ability, emphasizing that the use of interpretable models does not necessarily imply a reduction in performance. In the future, considering the positive results obtained with non-expert supervision, it would be worth studying the effects of expert supervision and gathering medical opinions on the clinical relevance and quality of prototypes, and consequently, on the explanation.

%% file: Sections/material_extra.tex
\section*{Supplementary Material}
\beginsupplement

\begin{figure}[th]
\centering
\includegraphics[width=\textwidth]{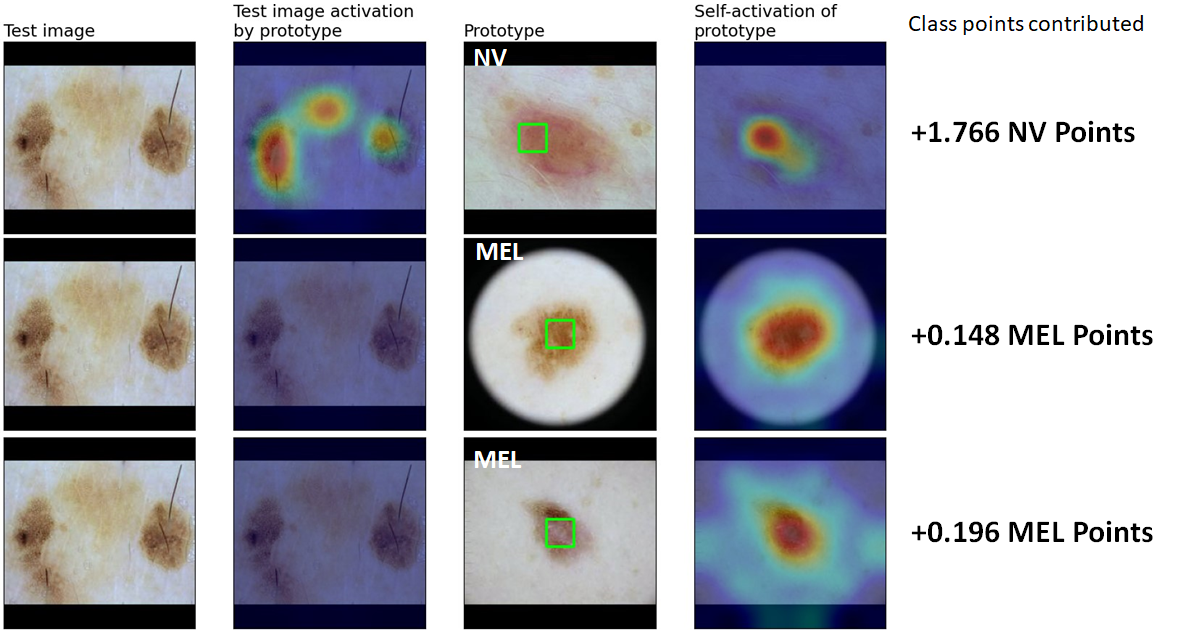}
\caption{Explanation of $L_\text{P}$ + $L_\text{M}$ approach for a validation case from the ISIC 2019 dataset with the top 3 activated prototypes in Densenet-169 backbone. The first prototype is nevus, while the other two are melanoma. The test image is from the nevus class and is correctly classified. We are only displaying 3 prototypes instead of the 18 prototypes, as a means of simplification. It's crucial to state that the total points assigned to the predicted class, across all its prototypes, exceed those of the other class. The substantial point difference between the top two prototypes is notable. The closest resembling prototype heavily influences the decision. } \label{}
\end{figure}

\begin{figure}[th]
\centering
\includegraphics[width=\textwidth]{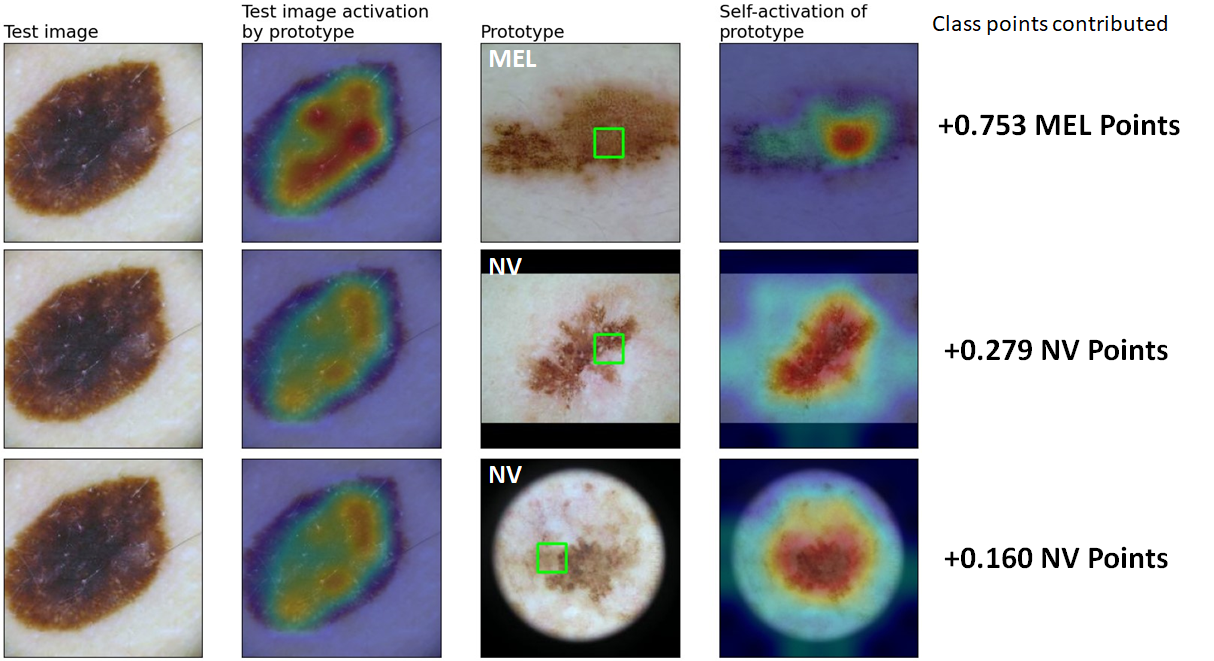}
\caption{Explanation of $L_\text{P}$ + $L_\text{M}$ approach for a validation case from the ISIC 2019 dataset with the top 3 activated prototypes in Densenet-169 backbone. The first prototype is melanoma, while the other two are nevus. The test image is from the melanoma class and is correctly classified. } \label{}
\end{figure}

\begin{table}[h]
\centering
\caption{Best hyperparameter configuration for the results in Table 1.}
\label{tab:hyp}
\begin{tabular}{|c|c|cc|}
\hline
\multirow{2}{*}{\textbf{Model}} & \multirow{2}{*}{\textbf{Approach}} & \multicolumn{2}{c|}{\textbf{Hyperparameters}} \\ \cline{3-4} 
 &  & \textit{\textbf{$D$}} & \textbf{top-\textit{k}} \\ \hline
\multirow{3}{*}{\textbf{ResNet-18}} & $L_{\text{P}}$ & 512 & 25 \\
 & $L_{\text{P}}+L_{\text{M}}$ & 256 & 28 \\
 & $L_{\text{P}}+L_{\text{R}}$ & 256 & 40 \\ \hline
\multirow{3}{*}{\textbf{ResNet-50}} & $L_{\text{P}}$ & 256 & 40 \\
 & $L_{\text{P}}+L_{\text{M}}$ & 512 & 13 \\
 & $L_{\text{P}}+L_{\text{R}}$ & 512 & 19 \\ \hline
\multirow{3}{*}{\textbf{EfficientNet B3}} & $L_{\text{P}}$ & 512 & 19 \\
 & $L_{\text{P}}+L_{\text{M}}$ & 512 & 13 \\
 & $L_{\text{P}}+L_{\text{R}}$ & 256 & 10 \\ \hline
\multirow{3}{*}{\textbf{Densenet-169}} & $L_{\text{P}}$ & 512 & 49 \\
 & $L_{\text{P}}+L_{\text{M}}$ & 256 & 31 \\
 & $L_{\text{P}}+L_{\text{R}}$ & 128 & 49 \\ \hline
\multirow{3}{*}{\textbf{VGG-16}} & $L_{\text{P}}$ & 128 & 112 \\
 & $L_{\text{P}}+L_{\text{M}}$ & 128 & 76 \\
 & $L_{\text{P}}+L_{\text{R}}$ & 128 & 112 \\ \hline
\end{tabular}
\end{table}